# Robust Boosting Forests with Richer Deep Feature Hierarchy

**Jianqiao Wangni**
Department of Computer and Information Science
University of Pennsylvania
Philadelphia, PA, USA
zjnqha@gmail.com

## Abstract

We propose a robust variant of boosting forest to the various adversarial defense methods, and apply it to enhance the robustness of the deep neural network. We retain the deep network architecture, weights, and middle layer features, then install gradient boosting forest to select the features from each layer of the deep network, and predict the target. For training each decision tree, we propose a novel conservative and greedy trade-off, with consideration for less misprediction instead of pure gain functions, therefore being suboptimal and conservative. We actively increase tree depth to remedy the accuracy with splits in more features, being more greedy in growing tree depth. We propose a new task on 3D face model, whose robustness has not been carefully studied, despite the great security and privacy concerns related to face analytics. We tried a simple attack method on a pure convolutional neural network (CNN) face shape estimator, making it degenerate to only output average face shape with invisible perturbation. Our conservative-greedy boosting forest (CGBF) on face landmark datasets showed a great improvement over original pure CNN methods under the adversarial attacks.

## 1 Introduction

In the past decade, deep learning has improved the performance on various computer vision tasks, including 3D face alignment [36, 44]and 2D face alignment [7, 14, 17]. Meanwhile, deep learning is also proven of alarming weakness against some adversarial manipulation in extreme cases [1]. By solving a reverse optimization problem with L-BFGS [33], or more efficiently, modified L-BFGS for non-smooth objectives [2, 37], or Fast Gradient Sign Method (FGSM) [19], a meaningful prediction could be easily turned to randomness. Those attacks not only disabled high-level semantic classifiers such as for ImageNet, to mispredict a cat to be a dog, an automobile to an airplane, some errors that are easily detectable by humans, the attacks also work for low-level tasks such as segmentation [21] and 3D transformed image recognition [4], which is an alarming signal for 3D reconstruction which are sensitive to pixel-wise features and targeting for geometric transformation. The deep feature hierarchy provides massive information, from low to high level, which combine together to ultra-high dimensional feature vectors, but the selection or screening procedure could be non-trivial, which partially makes most of the state-of-the-art framework to ignore lower level features for final prediction. We can certainly borrow inspiration from rich literature in this area [15, 28, 20, 34], but we can also use these rich features to improve robustness in one framework, such as using adversarial data as augmentation to derive better models [38].

We start from the fundamental advantage of deep networks over handcrafted features and shallow machine learning models: being *end-to-end trainable*. This helps with minimization of the loss during training, and also helps with maximization of the for searching adversarial examples. It inspires us to consider other approaches, such as decision trees, that have a discontinuity in output even with very small detail changes. The tree models are easy for going forward from features to predictions, but are difficult to go backward, which will be a combinatorial optimization, much harder than gradient-based approaches for neural networks [3] [11] [8]. So we try to compose a framework that uses both raw images and powerful semantic features from neural networks, and replace the prediction header that makes the final prediction using non-differentiable methods, or ensembles, e.g. boosting. In this paper, we will provide an orthogonal perspective to *enhance* the adversarial robustness, that being said, to do minimal effort on an existing algorithm, and could be combined with different methods from vision, graphics, and learning perspectives to improve robustness while retaining the accuracy.

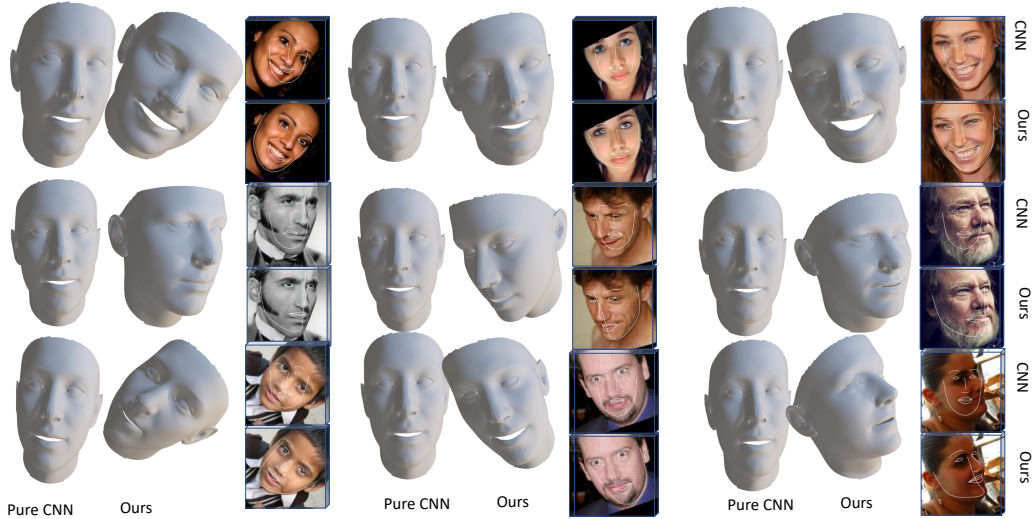

Figure 1: We show that even with less than $1\%$ pixel adversarial perturbation which is almost invisible to human eyes, it may lead to a failed 3D face reconstruction using pure deep learning method, and degenerate to average face shape, with only a slight bit of orientation information preserved. We propose an enhanced boosting method CGBF to easily plug in the same neural network and weights, without any tuning the network, and get a much more robust result.

## 2 Decision Tree for Adversarial Robustness

The Convolutional Neural Network (CNN) denoted as $F$, with input image $I$, target $y$, and loss function $L$, can be attacked by maximizing the loss function above as [19]

$$I' \leftarrow Clip[I' + \epsilon(\nabla_I \mathcal{L}(F(I'), y)] \qquad (1)$$

with network $F$ being fixed, where the clipping function enforces the perturbed image $I'$ to be valid. The attack takes advantage of the piecewise-linear region around an input example, and try to find a small perturbation along the gradient. We demonstrated in Figure.(14, ) that a small attack can make the pure CNN estimator output an average face, visualized in 2D sparse landmarks, depth, 3D dense landmarks, 3D orientation, projected normalized coordinate code (PNCC), which normalize the $3D$ coordinates, project to 2D and then color with RGB channels.

To defend this, we try to pursue a regressor that is insensitive to such perturbation, e.g. being piecewise constant within $\epsilon$ radius of input $I$, so that perturbation will not change the prediction. This naturally encourages us to take the decision tree model and enforce the samples $\{I^n\}$ to have a large enough margin around a split boundary. We use $X$ to denote a collection of training data. A decision tree has a structured series of decision nodes, where each node has a split function that partition the data space into two sides, by selecting a feature $d$ from all coordinates, and then applying a comparison of its value between the threshold $\tau$, and divides $X$ into left partition $X_L$ and right partition $X_R$, two of which will be treated independently and recursively split with other features:

$$Left: \quad X_L = \bigcup x | x_d < \tau, \quad Right: \quad X_R = \bigcup x | x_d \geq \tau, \qquad (2)$$

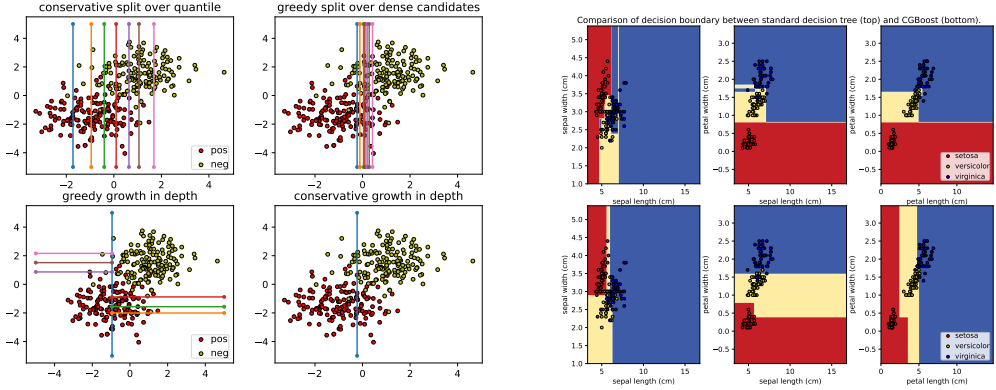

Figure 2: Comparison between different strategies for trees on Iris classification dataset. The upper row shows a standard tree with a maximum depth of 5 and maximum leaves of 32. The bottom row shows CGBF of 3 trees, having $4, 8, 12$ bins/leaves accordingly, and even combined together is still smaller than that decision tree. CGBF can find the proper dimensions and threshold to split to mimic optimal splits, e.g. several inclined lines on the 2D subspace.

and if $X_L, X_R$ are leaf nodes, the tree assigns values $\phi_{d,\tau}(x)$ that minimize internal losses accordingly. We use $S(X)$ to define the score function on data within $X$ if all data in $X$ are in one partition, or the purity of a data partition. The quality of a split can be defined by a score function $S(d, \tau)$, which by definition depends on the selected feature $d$ and threshold $\tau$. Suppose that $X_L$ and $X_R$ are two nodes after the split, we can also denote the score function as $S(X_L, X_R)$.

$$S(d, \tau) = S(X_L, X_R) = S(X_L) + S(X_R) - S(X), \tag{3}$$

The score function can be defined specifically for each task. For example, for a binary classification task where the score function is information gain so that $S$ represents the negative information entropy. For regression tasks, the score function typically depends on the Taylor expansion on the current estimate, which we will introduce later. We construct a decision tree that hopefully has less **controversial samples**: those predicted wrong since they fall within radius $\epsilon$ from threshold $\tau$ on the other side:

$$\min_{d,\tau} \sum_x \mathbb{I}(\|x_d - \tau\| < \epsilon, \phi(x) \neq y). \tag{4}$$

To help with the later illustration of our framework, we use two components to jointly represent a deep neural network: the layers to extract nonlinear representative features $F^{feature}$ and the header that locates $m$ points using these features. The two functions cascade together equal to the neural network, and will be trained as one network. We use $n$ to denote the image index out of $N$ images in total, and train the network as standard approach. We propose to use a cascade feature, that combines raw image pixels, the lower and middle layers of the deep network as

$$\hat{y}^n = F(x^n), \quad x^n = Concat[I^n, F^{feature}(I^n)], \tag{5}$$

and we will use $y^n$ to denote the target. We will describe the performance of using different layer of CNN in experiments, as this trigger a problem to a balance the pros and cons, since different layers have different sensitivity to the attack, so we need insensitive features $\{d\}$ for the task, and on the other hand, they contribute to the accuracy differently. We can also ensemble more layers to boost the performance but that might introduce additional memory overhead for storing CNN middle layers. Based on the score function in Eq.(3), we add a $\gamma > 0$ weighted penalty on adversarial attack gradients,

$$S(d, \tau) = S(X_L, X_R) - \gamma \mathbb{E}[\nabla_d \mathcal{L}(F(I), y)]. \tag{6}$$

The robustness of the decision tree also comes from the hardness to attack the cascade features. According to the research on attacking a single tree [29], since the leaf node has a piecewise constant output and has no informative gradient within the partition, it takes a complete traverse of decision nodes to find the optimal assignment, which could be done in $O((\#nodes)\log(\#nodes))$ time by efficient tree search and Mixed-Integer Linear Programming [25, 3]. Then to find a perturbation $I'$ to generate CNN features, to satisfy this target of attacking trees, are two contradicting objectives for two different layers. Based on the tree model, we naturally extend it to Gradient Boosting Decision Tree (GBDT) [18, 23, 12, 26], which is an ensemble method that combines the predictions over multiple trees. GBDT is trained in a stage-wise growing manner that each stage, a new tree is trained to fit the residual, from the current estimate of all previously trained trees to targets. Put GBDT in simplification:

$$\phi_K \leftarrow \nabla_F \mathcal{L}(F_{K-1}(I), y), \quad F_{K-1}(x) = \sum_{k=0}^{K-1} \phi_k(x), \qquad (7)$$

where $\phi_k$ is the $k$-th tree model and $F_K$ is first $K$ stage ensemble. To reduce model complexity, sometimes it applies a regularization term $\Omega$ to penalize the tree depth and the $\ell_2$ norm of the leave weights. GBDT is similar to the coarse-to-fine scale search ideas for facial landmarks: the data $x$ are image local features, like shape indexed features, with SIFT [40, 42] or Local Binary Features (LBF) [31], that are extracted around the current prediction of the landmarks. To construct the $K$-th tree, we need the gradient $g(x)$, and Hessian information $h(x)$ on $F_{K-1}$ if also available, the loss function is approximated with Taylor expansion as $l(y, \phi_K(x)) \approx l(y, \phi_{K-1}(x)) + g(x)\phi_K(x) + 1/2h(x)(\phi_K(x))^2$ where $\phi_K(x) = \sum_{k=1}^{K} f_k(x)$, and $f_K$ is optimized with regard to the function above. If the Hessian information is incorporated, e.g. like XGBoost [12], with regularization parameter $\lambda > 0$, the score function for regression tasks is $S(X) = -(\sum_{x \in X} g(x))^2 / \sum_{x \in X} h(x) + \lambda$.

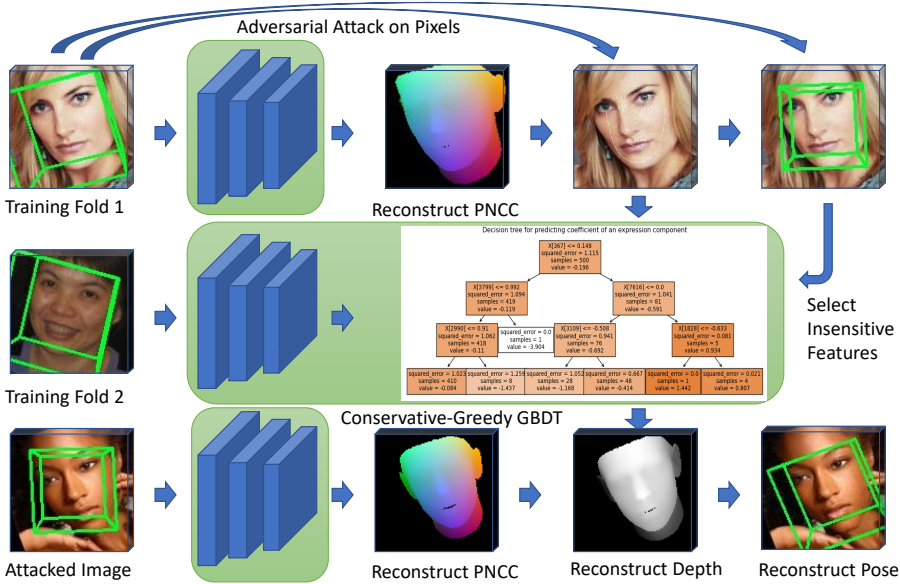

Figure 3: The overall algorithm framework. 3D face shape predicted by CNN reconstructs PNCC, 2D sparse landmarks, and oriented bounding boxes (not necessary in this order). The adversarial gradient is informative feedback to select stable features.

## 2.1 Greedy Growth in Depth and Conservative Splits

Although GBDT is composed of mostly non-differentiable functions, therefore, is much harder to attack, the vulnerability remains. We address the issue from several aspects. First, we use features extracted from different layers of deep networks, which cover both fine grain and coarse grain local features. Second, we encourage selecting robust features that are insensitive to perturbations. We separate the training pipeline to different data, to avoid searching the threshold by overfitting over fixed samples. To this end, we present the Conservative-Greedy Boosting Forest (CGBF) in the following, with a diagram in Figure.(3).

Our key insight here is: to adapt the gradient boosting forest for 3D face modeling, or image-based tasks in general, by combining features across layers we got ultra high dimensional features, lots of which are interchangeable with others, then it is not necessary to find an optimal split for each node.

A classical algorithm searching for a split would iterate all dimensions and all possible thresholds to make the best split, $d^\star, \tau^\star = \arg\max_{d,\tau} S(d,\tau)$, as a greedy selection. Since we need to balance this and the objective in Eq.(4), then we have to make less greedy progress in each split to avoid the controversial region of radius $\epsilon$, if many samples tend to be predicted wrong in this space. As we mentioned, we do not pursue that each split makes two children nodes, two partitions, to be both improved. we will also make a split that as long one child node is improved, therefore using this splitting objective function:

$$S(d,\tau) = \max[S(X_L, X_R), \beta \frac{|X_L|}{|X|} S(X_L), \beta \frac{|X_R|}{|X|} S(X_R)], \tag{8}$$

where $\beta \in [1, +\infty)$ is a tunable parameter. We take the split if one of $X_L, X_R$ is purified enough, and we leave the other child node to be split in the future, based on another different feature $d'$. The intuition behind this is that loss functions on sets are submodular in terms of partitioning: to put more and more samples from a set to its subset, the marginal gain will decrease each time, and the ambiguity of the subset will increase. We demonstrate in Figure.(2). To avoid mispredicting too much in the vertical split in the top row, it might be better to do a suboptimal split and then grow tree depth, to increase another split in the horizontal direction. Based on this, we will no longer regularize the depth of the tree, nor require the structure to be a balanced binary tree, since each split is conservative, and we need deeper tree depth to make it even. We use regularization $\Omega(f)$ that only relates to the leaf nodes,

$$\Omega(f) = +\infty \times \mathbb{I}(\#\text{leaves}(f) \leq \#\text{max leaves}), \tag{9}$$

which doesn't penalize any leaf node until reaching a maximum number. But we will set the number to be a power of 2 to simplify implementation. Note that our framework is motivated and adapted to imaging-related tasks, under the premise of having oversaturated features extracted by CNN and they are more interchangeable. This technique may not be suited to tasks in other areas such as information retrieval: e.g. to classify a webpage is the official company webpage or not, there are some important features like the number of webpages linking to it, and if this feature is handled and split well, it may be hard to remedy the result from other features.

## 2.2 Conservative Split

One heuristic way to avoid a controversial split is to further test an adjustment $\Delta\tau$ to the threshold $\tau^\star$, to see if it is better or worse. The controversy can be related to the number of data samples sacrificed for this optimal split, although the exact number might be super time-consuming to know. [10] choose to search over a much smaller set, to change an additional one or two samples $\Delta X$ near the decision border $\tau$ from left node to right node, or vice versa. We will refer to $\Delta X$ as the controversial data, and $\Delta X_L, \Delta X_R$ are within $X_L, X_R$ accordingly. So this derives three more cases: to put all samples in $\Delta X$ in $X_L$, or to $X_R$, or to swap:

$$S_1 = S(X_L, X_R), \quad S2 = S(X_L \cup \Delta X_R, X_R \cup \Delta X_L) \tag{10}$$

$$S_3 = S(X_L, X_R \cup \Delta X), \quad S_4 = S(X_L \cup \Delta X, X_R), \tag{11}$$

and the controversy is defined over the best case versus worst case in the controversial space

$$CTV(d,\tau) = \max(S_1, S_2, S_3, S_4) - \min(S_1, S_2, S_3, S_4). \tag{12}$$

We take a step further. By subtracting this term from the score function, we got an objective function to maximize the lowest score within $\Delta\tau$. We increase $\Delta\tau$ or equivalently $\Delta X$, to be a fixed percentage of the training data, e.g. $1/32$ of the overall data, then the worst case can be approximated by including or excluding this $1/32$ portion of data to $X_L$, which narrows down the search procedure to the quantile of ranked data, with $X_L$ taking $1/32, 2/32, 3/32, \cdots$ of $X$. To this objective, we put data samples into equal-sized bins according to $d$-th feature. Suppose we have $J$ bins, $X_j$ denotes the samples in the bin $j$, and $\tau_j$ is the threshold to distinguish $X_j$ and $X_{j+1}$, then optimize over $J$ possible splits. Starts with

$$\forall j, |X_j| = \frac{1}{J}|x|, \quad \text{start:} X_L = \{\}, X_R = X, \tag{13}$$

and repeat the following procedure

$$S(d,\tau) = \max_{d,\tau}\{S(X_L \cup X_j, X_R), S(X_L \cup X_j, X_R)\}, X_L \leftarrow X_L \cup X_j, \quad X_R \leftarrow X_R \setminus X_j.$$

The quantile search above has different intuition from XGBosst and LightGBM [12, 26] which mainly consider computation efficiency. Here our approach considers it to be an important robustness contributor, as it is the main technique for our *Coarse-to-Fine Splitting* motivation that gradually prioritizes accuracy over robustness as the learning rate decreases, carefully adjusting the granular. We construct trees from the first one with fewer bins and tree leaves, each bin consisting of $1/16, 1/32$ of data, to the last tree that each bin having $1/256, 1/512$ data. In Figure.(2) we demonstrate the Iris dataset with 3 features for classification, with the top row consisting of one tree of greedy search and more leaf nodes, there is a tendency to have a small gap area or slices generated from overfitting, vulnerable to perturbation.

---

**Algorithm 1** Enhancing Robustness of Deep Networks for 3D Face Models

---

**Require:** Trained neural networks $F$ and defining two subnetworks as $F^{Feat}, F^{header}$.
   Extract features $F^{Feat}(x)$ from training folds $X_{optfeat}, X_{optthresh}, X_{optpred}$.
   Set #(max-trees). Start with tree setting #(max-bins) = 32, and #(max-leaves) = 32.
   **while** K (#trees) $\leq$ #(max-trees) **do**
      **while** #(leaves) $\leq$ #(max-leaves) **do**
         Calculate the gradient $g(x)$ and Hessian $h(x)$ of $F_k(x) = \sum_{k}^{K-1} \phi_k(x)$ on $X_{optfeat}$.
         Divide the features $\{F^{Feat}(x)|x \in X_{optfeat}\}$ into #(max-bins) bins, select the top $D$ dimensions to split according to Eq.(16).
      **while** #(leaves) $\leq$ #(max-leaves) **do**
         Calculate $g(x), h(x)$ on $X_{optthresh}$. Divide the features $\{F^{Feat}(x)\}$ into bins,
         Select the optimal threshold $\tau$ on feature $d$ to split according to Eq.(17).
      **while** #(leaves) $\leq$ #(max-leaves) **do**
         Calculate $g(x), h(x)$ on $\sum_{k}^{K-1} \phi_k(x)$ with training fold $X_{optpred}$. Assign data to leaves of trees built above, and assign the prediction value of leaves.
      Add the fitted tree $\phi_K$ to the forest.
      If not making enough progress, increase #(max-bins), #(max-leaves).

---

### 2.3 Cross Training for Individual Phase

The classical method of training gradient boosting machines have the shared objective function for entire pipeline for training tree: 1), OptFeat: searching for optimal features $\{d\}$ to split,
2), OptThres: searching for the optimal threshold $\{\tau\}$ to split based on $\{d\}$,
3), OptPred: the prediction value $\hat{y}$ for the data falling into the leaf nodes based on $\{d, \tau\}$.
This is straightforward to understand, but exposes the model to adversarial attack by overfitting, if the gradient information were to be leaked, since overfitting generates a controversial area of more misprediction cases. We try to avoid this, by importing the idea of bagging and cross-validation to the training phase: that we divide and distribute the pipeline to be accomplished on different data, as in Figure.(3). Suppose we find a bad split accidentally, by the greedy nature, but the split varies a lot driven by lots of controversial samples nearby, the threshold will be drastically different in another sample set. So borrowing a threshold from other data will avoid this one and just include more controversial samples on one side, $X_L$ or $X_R$, but not across them. This will make conservative progress in fitting and signals a future stage tree to fit a proper residual. We refer to this as cross-training. We divide the training set $X_{train}$ into several folds, say $X_{optfeat}, X_{optthres}, X_{optpred}$ and they form the original set together

$$X_{optfeat} \cup X_{optthres} \cup X_{optpred} = X_{train}, \qquad (14)$$

and they are supposed to be disjoint with each, although no need to exactly rigorous, if the joint set is smaller in magnitude,

$$|X_{train}| \gg |X_{optfeat} \cap X_{optthres} \cap X_{optpred}| \qquad (15)$$

and we then perform the procedures above using different folds. First, we start with selecting optimal feature $d$ as

$$d^\star = \arg\min_d \mathbb{E}_X \|\nabla_d F^{header}(x, y)\|^1 \quad | \quad X = X_{optfeat}, \qquad (16)$$

note that $x = concat[I, F^{feature}(I)]$, and the gradients for both middle layer and raw pixels can be get via one round of backpropagation. Next is to determine the threshold to split,

$$\tau^\star = \arg\min_\tau \mathbb{E}_X S(d^\star, \tau), \quad | \quad X = X_{optfeat}. \tag{17}$$

Suppose after recursive splitting at $d^\star, \tau^\star$, we got a subset on leaf node $X_{leaf} \subset X_{optpred}$, then their prediction value will be

$$\phi(x) = \sum_{x \in X_{leaf}} g(x) / (2 \sum_{x \in X_{leaf}} h(x)), \quad \forall x \in X_{leaf}, \tag{18}$$

where $h, g$ are gradient and Hessian respectively.

## 3  Application in 3D Face Modeling

Face analytics has been widely used with security concerns, therefore need more robust models. To recognize a person from images, there are several steps necessary in general: first to locate the bounding boxes of faces, usually described by a four dimension vector such as the width, height, and center of the box, and then locate the sparse facial landmarks, typically of 68, e.g. at the corners or perimeters or eyes, nose, eyebrows, and facial skeleton [32]. The facial region, or region of interest (ROI) will be cropped out, resized and transformed to a standard pose, and then performed recognition procedure. More accurate landmarks give better geometric transformation and thus a better recognition result [41], and there are some works to theoretically prove the estimation accuracy [39]. 3D reconstruction of tens of thousands of dense facial landmarks, provides richer geometric information, which is crucial to the entertainment industry such as computer animation [9]. The animation character can be manipulated based on the facial landmark variation of human actors. The 3D geometric features are capable of recognizing a person [30], and being intrinsically invariant to head pose and lighting conditions. The 3D reconstruction can be used in face identity forgery for financial fraud and used against such forgery [45].

One type of facial shape detection regards the problem as multivariate regression for sparse 2D points, i.e. detecting 68 landmarks is treated as 136-dimensional multivariate regression from high dimensional local features [40, 42, 31]. Another way is to impose the linear subspace constraint on the landmarks, like the active shape model (ASM) [13], so the shape variations are within the predefined face appearances, which can be extracted by principal component analysis. These algorithms can be generalized to 3D landmarks, like 3D Morphable Models (3DMM) [6]. By sharing model parameters with different numbers of landmarks, we can alternately estimate the dense and sparse shapes, one from another. By aligning the linear combination coefficients of 2D and 3D shapes, or registering 3D landmarks with their 2D projection, we can infer a 3D shape from 2D facial images. So in our paper, we will focus on estimating the common shape parameters that reconstruct both 3D/2D-dense/sparse face shapes.

The direct learning approach to face alignment can be very simple and straightforward. Taking the input image $I$ and annotation of $m$ 2D/3D face landmarks, which stack together and get a shape vector $y^{2D} \in \mathbb{R}^{2 \times m}$ or $y^{3D} \in \mathbb{R}^{3 \times m}$, including 2D/3D location for each landmark. A dense 3D face model consists of a mesh with thousands of vertices and triangles, therefore is difficult to directly construct from a 2D image. Following recent works [24] [16] [35] built on 3D Morphable Model (3DMM), each 3D face can be seen as a standard average face shape plus a linear combination of some representative face shapes, which either are related to their identity or to their expression. We denote the average face $\bar{Y} \in \mathbb{R}^{3 \times m}$, and the 40 identity components $A_{id} \in \mathbb{R}^{3 \times m \times 40}$ and 10 expression components $A_{exp} \in \mathbb{R}^{3 \times m \times 10}$, where each component can be get from principal component analysis (PCA). Both 2D and 3D shapes can be expressed as

$$y^{2D,3D} = \bar{y}^{2D,3D} + A_{id}^{2D,3D} \alpha_{id} + A_{exp}^{2D,3D} \alpha_{exp}, \tag{19}$$

where the corresponding coefficients $\alpha_{id} \in \mathbb{R}^{40}$ and $\alpha_{exp} \in \mathbb{R}^{10}$ are shared by both 2D and 3D model. The 2D shape can be get via orthographic projection from 3D,

$$y^{2D} \propto Proj \times R \times y^{3D} + t \tag{20}$$

where $Proj$ is the 3D-2D projection matrix, $R \in \mathbb{R}^{3 \times 3}$ is the rotation matrix of the object, and $t \in \mathbb{R}^2$ represents the 2D translation, and $\propto$ omits a scale factor dependent on camera focal length.

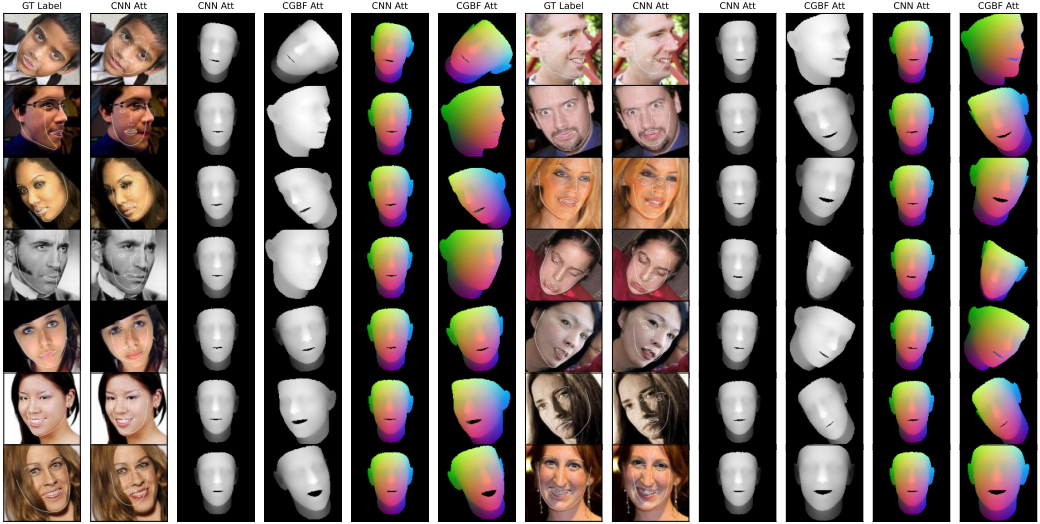

Figure 4: Comparison of 2D sparse landmarks (with ground truth GT label), depth, and PNCC (normalized dense 3D coordinates) reconstruction by the pure CNN and CGBF on attacked images.

So if the coefficients $\alpha$ can be inferenced from sparse landmarks on 2D images, then the 3D model with both sparse and dense landmarks can be all built based on the same coefficients $\alpha$.

To estimate the shape $y$ from plain 2D image $I$, we can certainly design a deep neural network to directly predict the value of $y$, or indirectly predict $\alpha$ and geometric transformation $R, t$.

$$F(I) \to y^{2D,3D}, \quad Or, \quad F(I) \to \alpha, R, t. \tag{21}$$

There are some possible loss function for neural network training, e.g. parameter distance cost (PDC), vertex distance cost (VDC), and weighted parameter distance cost (WPDC),

$$\mathcal{L}_{PDC} = \|concat[\hat{\alpha}, \hat{R}, \hat{t}] - concat[\alpha, R, t]\|^2 \tag{22}$$

where $(\hat{\cdot})$ represents the prediction from neural network or other models; $VDC$ loss takes all points into account and $WPDC$ takes additional importance vector $w$,

$$\mathcal{L}_{VDC} = \|\hat{y} - y\|^2, \quad \mathcal{L}_{WPDC} = (\hat{y} - y)^T \text{diag}(w)(\hat{y} - y). \tag{23}$$

and in implementation of [44], the weight $w$ is related to the degradation error caused by each coordinate if change it from ground truth to a prediction value.

## 4 Experiments

We build our framework on top of the implementation and datasets designed in 3DDFA [44]. 300W [32] dataset combines multiple face alignment datasets such as AFW [43](5,207 samples), LFPW [5] (16,556 samples), HELEN [27] (37,676 samples). 300W-LP built on 300W with a technique of reverse face frontalization, to generate augmented images with different 3D transformations, although it has a loss in realisticity. 300W-LP is divided into 636,252 training images and 51,603 validation images. We implemented the CGBF algorithm with Scikit-Learn. We pick two layers from the neural networks, and take 4096-dimensional features from each layer. The features are selected by gradient $\ell_1$-norm from the adversarial attacks. The target $y$ has 62 elements, with 50 elements representing the coefficients for face components, 9 elements for the rotation matrix, and 3 for the translation. To build the gradient boosting forests, we set the maximum number of trees to be 40 for each target dimension. We set the maximum bins and leaves to be 32 to start, and increase to 64 after 30 stages. During experiments, we found that the geometric transformation components $R, t$ are easier to fit, but some components like expression components $\alpha_{exp}$ are more difficult to fit in the sense that training loss decreases much faster than validation loss. We then found the best practice for fitting is to set the learning rate for all coordinate of $y$ to be 0.3, and the loss on different components decrease at different rates. For the cross-training purpose we described, in the experiments, we divide the training set into folds, each of which contains a maximum of 32000 images, and distribute CGBF training pipeline on each fold.

According to the result [44] that PWDC is empirically the best single metric among the three, and PWDC plus VDC are empirically the best if mixture loss is used. We will use both PWDC and VDC loss for generating adversarial examples with FGSM, and will report all three loss metrics in the result. We tried using PWDC and VDC for measuring feature robustness in Eq.(6), and it turns out PWDC is slightly better, so we report our result with PWDC gradients as robustness measurement. We tested the framework performance variation by using different layers of the neural networks, plus using the pixels of the raw image, which will be noted as layer $0$. The neural network architecture inherits the setting to use MobileNetV1 [22] that has up to 15 blocks of convolutional layers. We tested the performance of pure boosting forest (CGBF), pure CNN, and a mixture model that linearly combines the prediction of both models with different weights like $(0.3, 0.7), (0.5, 0.5), (0.7, 0.3)$, and got the results in between the pure CGBF accuracy and the pure CNN accuracy. We tested the performance under different attack step sizes as well. As the attacks become stronger, the reconstruction loss of CGBF increases alongside pure CNN, and this should be the fundamental limitation of the method as it uses CNN features and cannot fully avoid perturbation negative impact. The results for same setting may have differences from random data loader and batching.

Table 1: Performance on 300W-LP validation images $I$ and attacked image $I'$. CGBF extracts features from two layers from CNN, compares with CNN under different loss metric and attack sizes (with image pixels normalized to [0,1]).

| Attack size | | $\|I - I'\| \leq 5e-3$ | | | $\|I - I'\| \leq 1e-2$ | |
|---|---|---|---|---|---|---|
| | $F^{CNN}$ | $F^{CNN}(I')$ | $F^{CGBF}(I')$ | | $F^{CNN}(I')$ | $F^{CGBF}(I')$ |
| layer[0,12] | improved | | | | | |
| VDC | 3494.52 | 5335.24 | 4122.71 | | 8069.29 | 5548.10 |
| WPDC | 4.95 | 6.57 | 4.99 | | 8.05 | 5.38 |
| PDC | 72.05 | 81.81 | 74.70 | | 89.53 | 78.40 |
| layer[4,12] | strongly | improved | | | | |
| VDC | 3494.52 | 5322.63 | 3619.52 | | 8069.29 | 4948.92 |
| WPDC | 4.95 | 6.56 | 4.91 | | 8.06 | 5.23 |
| PDC | 72.05 | 81.81 | 70.79 | | 89.53 | 73.92 |
| layer[6,12] | strongly | improved | | | | |
| VDC | 3494.52 | 5351.28 | 3649.67 | | 8069.29 | 4992.27 |
| WPDC | 4.95 | 6.57 | 4.88 | | 8.05 | 5.20 |
| PDC | 72.05 | 81.81 | 70.63 | | 89.53 | 73.74 |

We can see an interesting patterns related to layer-depth from Table.(1), with all numbers times by $\times 100$ to be standardized. Since the deeper layers change more drastically under adversarial perturbation, shallow layers and input images are more stable due to the vanishing gradient phenomenon during backpropagation. On the other hand, deeper layers are more expressive and help with better accuracy. Out of 15 blocks of convolution layers, the deepest layer $15$-th is not suited for training boosting forests comparing to $12$-th layer: this layer outputs a high-channel low-resolution feature map of $4 \times 4 \times 1024$, which greatly loses localized information that helps this task. Comparing the result of using layers $[0, 12]$ against $[4, 12]$ and $[6, 12]$ we found that layers 4,6 strike better balance than layer 0, i.e. the input pixels: each feature has a suitable size of receptive fields, not too localized nor globalized; they are relatively stable for the trees trained with conservative splits.

## 5 Conclusion

In the paper, we presented a minimal effort plugin to enhance the robustness by the boosting forest with a conservative-greedy balance to avoid controversial splits. This approach is orthogonal to existing machine learning research on the robustness of CNN, and vision/graphics research to incorporate better geometric and photometric cues to defend against worst cases. We experimented the robustness of deep learning regression approaches to 3D face reconstruction and proved our framework effective.

## 6 Acknowledgement

Thanks for someone to whom I cannot refer explicitly by name here, for insightful discussion, priceless encouragement, and incredible value of support from all other aspects, through my tough times, and shined light in pure darkness surrounding me.